\documentclass{sigkddExp}
\usepackage{graphicx}
\usepackage[para,online,flushleft]{threeparttable}
\usepackage[colorlinks, citecolor=red]{hyperref}

\begin{document}
%

\title{Monitoring Chinese Population Migration in Consecutive Weekly Basis from Intra-city scale to Inter-province scale by Didi's Bigdata}
\numberofauthors{1}

\author{
%
\alignauthor Renyu Zhao \\
       \affaddr{Didi Labs}\\
       \affaddr{Beijing, P.R.China}\\
       \email{zhaorenyu@didichuxing.com}
\alignauthor G\\
       \affaddr{Institute}\\
       \affaddr{x}\\
       \email{x}
}
\date{29 March 2016}
\maketitle
\begin{abstract}
Population migration is valuable information which leads to proper decision in urban-planning strategy, massive investment, and many other fields. For instance, inter-city migration is a posterior evidence to see if the government's constrain of population works, and inter-community immigration might be a prior evidence of real estate price hike. With timely data, it is also impossible to compare which city is more favorable for the people, suppose the cities release different new regulations, we could also compare the customers of different real estate development groups, where they come from, where they probably will go. Unfortunately these data was not available.

In this paper, leveraging the data generated by positioning team in Didi, we propose a novel approach that timely monitoring population migration from community scale to provincial scale. Migration can be detected as soon as in a week. It could be faster, the setting of a week is for statistical purpose. A monitoring system is developed, then applied nation wide in China, some observations derived from the system will be presented in this paper.

This new method of migration perception is origin from the insight that nowadays people mostly moving with their personal Access Point (AP), also known as WiFi hotspot. Assume that the ratio of AP moving to the migration of population is constant, analysis of comparative population migration would be feasible. More exact quantitative research would also be done with few sample research and model regression.

The procedures of processing data includes many steps: eliminating the impact of pseudo-migration AP, for instance pocket WiFi, and second-hand traded router; distinguishing moving of population with moving of companies; identifying shifting of AP by the finger print clusters, etc..
\end{abstract}

\begin{keywords}
Population Migration, Big Data, Access Point, Machine Learning
\end{keywords}

\section{Introduction}
Migration plays key role in the complex process of civilization and globalization. From states to regions, it impacts the polities and economics. With a grasp of migration in China for instance, one would be better understanding what is going on along with the process of rapid civilization, and how the policies enacted by the governments works or why they does not work\cite{chinamove}\cite{chinaubb}. There are many interesting studies on population migration in China, some compared intra-provincial migration with inter-provincial migration\cite{intraprovince}, some exploited into the full picture of population migration at provincial level\cite{interprovince}\cite{provincial}, some investigated in people moving within the city \cite{intracity}.

These studies draw a vivid picture on the process of people moving from one place to another in China, trying to explain and to exploit those behind the phenomenon, which are helpful in understanding what happened and what is happening in China. However, what they are in common is that the data they analysis are exclusively from the national census taken by the State Statistical Bureau of China, the census is carried out almost every 10 years, therefore makes the analysis dated and not be able to catch some new phenomenons in this rapidly changed period, not mention that it would consume dramatic resources to conduct such an national census. So far before we exploit Didi's big data, this detailed, but lagging data is the only source for demographer and urban planner to do researches\cite{updata}\cite{provincial}.

Recently Baidu revealed a heat map visualizes the Chinese New Year migration\footnote{http://qianxi.baidu.com/}, it is done by consecutively record position of Baidu Map's users, and process to put them together for demonstration\cite{baidu}. This feature is adept at demonstrating such real time massive migration, unfortunately when it comes to moving of permanent residential place, it lacks proper evidences. On one hand, all travelings and tourism are mixed in the data, the users might break the session, or even uninstall its map applications and switch to other, so that no more data is available. What's more, the social class distribution of users of Baidu Map product might varies from time to time, which makes the statistics unstable.

In this work, we propose an efficient and statistically stable way to percept and record population migration, and demonstrator some findings which are urban planners and investment decision makers might be interested.

\begin{figure}
\centerline{\includegraphics[width=8cm]{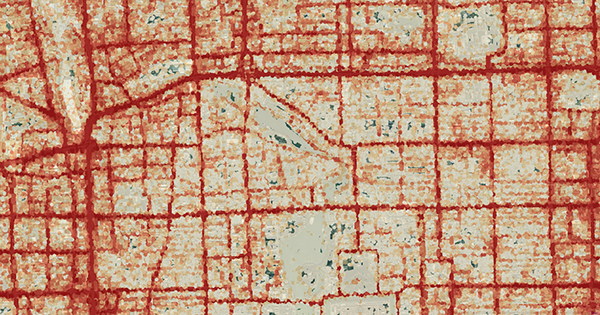}}
\caption{Demonstration of scanned WiFi signal density in north part of downtown Beijing}
\label{fig:bjsig}
\end{figure}

As aforementioned, people always move with their private AP nowadays, and the APs are universally detectable, therefore as long as we are able to scan WiFi signal universally every day, and distinguish home-set AP from others, the moving of people can be sensed. Figure \ref{fig:bjsig} shows WiFi signal density Didi scanned in a normal day, the deeper colored red, the more signals canned in corresponding area, mostly it is on the street, that is for the reason that open space is more appropriate for receiving signals. As almost all the area are covered on the map, most AP can be scanned.

The following section will be detailed introduction how we processing the data, and how the monitoring system works in Didi, then sections of exploited data followed, in the end some interesting future work will be introduced.

\section{How Didi's data built to the monitoring system}
\begin{figure}
\centerline{\includegraphics[width=8cm]{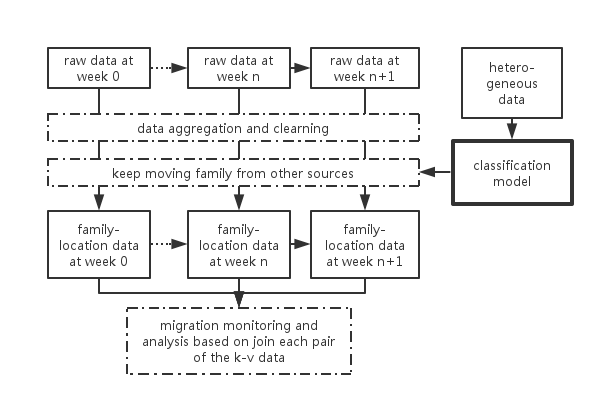}}
\caption{The framework of Didi's migration monitoring system}
\label{fig:system}
\end{figure}

The objective of this system is not only to monitoring the migration, but also to do data mining and more research in different aspects, so it is designed as open data API in the fundamental level. The system, as shown in Figure \ref{fig:system}, clean collected AP data to remove noises in the first place, then distinguish family AP from other sources with the help of heterogeneous data generated classification model, and yield data of family to location pairs. Afterwards in the analysis processed, for example, if population migration from January 2016 to March 2016 is in favor, the weekly data in these two months would be aggravated by consecutive right join operation, then a left join operation for the monthly data in time order.
\subsection{Remove noises}
Pure data cleaning is another topic in the field of data mining, it mainly resolve problems on data quality, we would leave it out in this paper. There are some more work for us when clean data is obtained.

One of such noises is second-hand trade, this activity also can be observed as AP moving from one place to another. Therefore we have observed major online second-hand trade platforms(2.taobao.com, www.ganji.com, www.58.com), which covers most of second-hand trade in China, it turns out router related transactions accumulated at a month basis takes less than 0.01\% of that in our system, therefore we regard it as not has trivial effect in the overall analysis, the real movement are far outnumbered these AP movements caused by second hand trade. When it comes to inter-city migration, second-hand AP trade is far more less. Keep in attention that there are some more exquisite analysis in special cases, where there are only few samples as a certain origin or destination of migration, in studying of these cases, the second hand trade should be paid attention to, so as to eliminate possible impact.

The major noises are from moving of companies, the moving of companies contribute a lot in the observation of AP migration. We deal with this problem with heterogeneous data generated classification model. There are many heterogeneous data ready and can be served as feature in the classification model. For instance, we are able to distinguish each AP from each building (in a probability distribution), in some cases even different floors of a building, then with the help of reverse-geocoding service, the properties of the building is get as features in the classification model.

\subsection{Model training}
In real practice, we made a Gradient Boosting Decision Tree (GBDT) classification model.
For the training set $\! \{(x_i, y_i)\}_{i=1}^n$, and a differentiable loss function $\! L(y, F(x))$, the gradient boosting method update model through aggregately fitting the pseudo-residuals $r$\cite{gbdt}:

$r_{im} = -\left[\frac{\partial L(y_i, F(x_i))}{\partial F(x_i)}\right]_{F(x)=F_{m-1}(x)} \quad \mbox{for } i=1,\ldots,n.$

at each step $m$, until converge.

The features are listed in Table \ref{tab:feature}, and the parameter of number of stumps is set as 4, based on cross validation experiments.
\begin{table}[h]
\caption{GBDT model feature and type}
\begin{center}
\begin{threeparttable}
\begin{tabular}{l c}
    \textbf{Feature description} & \textbf{type}  \\ \hline
    Property 1 type & nominal\\
    Property 1 probability & float\\
    ... & ...\\
    Property x type & nominal \\
    Property x probability & float \\
    history of connected terminal & int \\
    accumulated connected AP count & int \\
    simultaneous max connected AP count & int \\
    connected records day-night percentage & float \\ \hline
\end{tabular}
\begin{tablenotes}
\item[1] There for 4 types of property: 0-office building 1-residential building 2-mixture 3-uncertain;
\item[2] The ratio is defined as: Sum of Counts at (10:00-16:00)/Sum of Counts at (22:00-6:00)
\end{tablenotes}
\end{threeparttable}
\end{center}
\label{tab:feature}
\end{table}

The model is supervised model, and the target of this model is to determine whether the AP is residential or not.
Therefore a core work is to label some data for training the model. In our case we designed a method using user session to batch label data for training process. Suppose we designate residential AP as positive sample, and non-residential AP as negative sample, we split user session by time and location so as to identify some positive and negative samples. For example, if a AP is connected by a terminal which called a car service after 9 pm and leave in the next morning, it would be labeled positive, and if it is connected after 9 am till leave in the evening, it would be labeled negative.

Feature data are then aligned with the labels and randomly mixed. Top 2 possible buildings are selected for each AP (i.e. x=2 in the model feature table). In this way the training set $\! \{(x_i, y_i)\}_{i=1}^n$ is ready.

CART algorithm is employed to build stumps here, and a gradually decreased learning rate $\gamma$ is introduced as regularization term to avoid over-fitting.

The cross-validation result shows that the precision is around 97.0\%.

\subsection{AP print conformity checking algorithm}
Normally we judge if the AP has been moved by compare the conformity of its prints at the two time-stamps. The mostly used algorithms are Pearson Correlation Coefficient, Cosine Similarity and Mahalanobis Distance etc., some measures distance, which is counterpart os similarity, and some rectifies data for normalization purpose. In this case we turn the prints into simple vectors (e.g. $A$, $B$) and calculate cosine similarity, and in practice, most of the AP are moved far from original place, and the similarity naturally become zero, therefore the threshold is easy to set.
\begin{displaymath}\text{sim} = \cos(\theta) = {\mathbf{A} \cdot \mathbf{B} \over \|\mathbf{A}\| \|\mathbf{B}\|} = \frac{ \sum\limits_{i=1}^{n}{A_i  B_i} }{ \sqrt{\sum\limits_{i=1}^{n}{A_i^2}}  \sqrt{\sum\limits_{i=1}^{n}{B_i^2}} }\end{displaymath}

\section{Multi-scale migration observation}
In this section, we present several arranged results yield by data of migration between September and December in 2015, which is generated by the aforementioned monitoring system. More than a million migration are gathered in this time period.

\subsection{Inter-provincial and Intra-provincial migration}
There are some noticeable points in inter-provincial and intra-provincial migration:
\begin{itemize}
\item Most provinces facing net emigration, accumulation phenomenon is significant.
\item Magnitude between cities are interactive courses.
\item The amount of intra-province migration is also in conformity with economic activity of that area.
\item Beijing is special among the economic zones.
\end{itemize}

\begin{figure}
\centerline{\includegraphics[width=8cm]{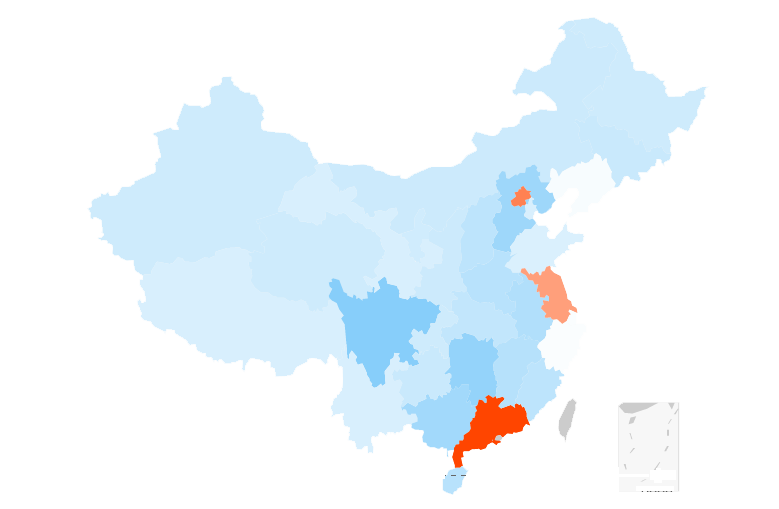}}
\caption{Net immigration density}
\label{fig:prio}
\end{figure}

Figure \ref{fig:prio} shows net immigration (immigration - emigration) density of each provinces, and top 6 and bottom 4 are regularized and listed in Table \ref{tab:prio}. As in the figure, the more dense in red, the more net immigration counts, the more blue, the more emigration counts.

\begin{table}[h]
\caption{Top 6 and bottom 4 provinces in net immigration}
\begin{center}
\begin{threeparttable}
\begin{tabular}{l c}
    \textbf{Province} & \textbf{Regularized net immigration count}  \\ \hline
    GUANGDONG & 1.00 \\
BEIJING & 0.69 \\
JIANGSU & 0.58 \\
ZHEJIANG & 0.22 \\
SHANGHAI & 0.20 \\
LIAONING & 0.19 \\
GUANGXI & -0.29 \\
HEBEI & -0.32 \\
HUNAN & -0.37 \\
SICHUAN & -0.51 \\ \hline
\end{tabular}
\end{threeparttable}
\end{center}
\label{tab:prio}
\end{table}
It is clear through observing the figure, that 20/80 law fits immigration/emigration as well as counts of migration in destination, it looks as if people from all provinces moving to the three provinces: Guangdong, Beijing and Jiangsu.

As to cities, Table \ref{tab:ctio} shows that it is the same situation as the cases of provincial data, 30.2\% of the cities are under positive net immigration, and the listed top 8 cities bears 72.4\% of all positive net immigration counts.

This table might be a perfect explanation of the rocket price hike of real estate in Shenzhen and Suzhou during last half year. All the other cities are far behind in regard of net population immigration.

As for Beijing, obviously through this table, it is clear people are still flow-in, which is against the urban plan enacted by the government. However this data is stationary, it is possible that the policies actually works, only slowly. In this case, a continuous observation is necessary, since if the policies of cutting down capital population works, we would see a modest decrease in net immigration count periodically. Next section would demonstrate the use of periodical monitoring data.
\begin{table}[h]
\caption{Top 8 and bottom 6 cities in net immigration}
\begin{center}
\begin{threeparttable}
\begin{tabular}{l c}
    \textbf{City} & \textbf{Regularized net immigration}  \\ \hline
    SHENZHEN & 1.00 \\
SUZHOU & 0.91 \\
BEIJING & 0.75 \\
CHENGDU & 0.46 \\
HANGZHOU & 0.40 \\
FOSHAN & 0.37 \\
WUHAN & 0.30 \\
SHANGHAI & 0.22 \\
NANPING & -0.08 \\
SHAOYANG & -0.08 \\
MEISHAN & -0.09 \\
ZIGONG & -0.09 \\
NEIJIANG & -0.13 \\
NANCHONG & -0.16 \\ \hline
\end{tabular}
\end{threeparttable}
\end{center}
\label{tab:ctio}
\end{table}

Although Beijing is in the second place of net immigration among all provinces and cities, the counts of immigration, emigration and intra-provincial migration are in the same magnitude, nearly larger by an order of magnitude compared with net immigration count. The destination people moving from Beijing and the origin from where people move to Beijing are also in conformity as shown in Table \ref{tab:bjio}.
\begin{table}[h]
\caption{Beijing's top 9 immigration source and top 9 emigration destination}
\begin{center}
\begin{threeparttable}
\begin{tabular}{l c l c}
    \textbf{Source} & \textbf{Cnt.} & \textbf{Destination} & \textbf{Cnt.} \\ \hline
    SHANGHAI & 1.00 & SHANGHAI & 1.00 \\
SHENZHEN & 0.71 & CHENGDU & 0.68 \\
CHENGDU & 0.70 & LANGFANG & 0.67 \\
TIANJIN & 0.63 & TIANJIN & 0.66 \\
HANGZHOU & 0.57 & SHENZHEN & 0.63 \\
SUZHOU & 0.54 & HANGZHOU & 0.60 \\
WUHAN & 0.51 & GUANGZHOU & 0.53 \\
LANGFANG & 0.48 & WUHAN & 0.50 \\
GUANGZHOU & 0.45 & SUZHOU & 0.42 \\ \hline
\end{tabular}
\end{threeparttable}
\end{center}
\label{tab:bjio}
\end{table}

\begin{figure}
\centerline{\includegraphics[width=8cm]{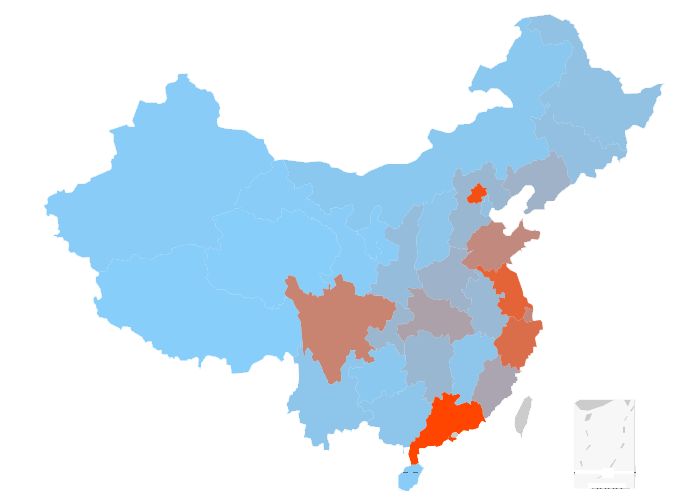}}
\caption{Intra-province migration density}
\label{fig:intropr}
\end{figure}

Intra-province migration could also reflect economic vitality, Figure \ref{fig:intropr} shows intra-province migration density.

If nearby cities are considered as an integrated party and combined together, there are three parties dominates in China: Beijing Area (Beijing and Tianjin in this paper), Yangtze River Delta (Shanghai, Suzhou, Hangzhou in this paper), and Pearl River Delta (Guangzhou, Shenzhen, Foshan, Dongguan in this paper). In Table \ref{tab:area} it is clearly shown that most migration happens in between these economic giants. Another significant phenomenon is that although the interactive within Beijing area is least active, the city of Beijing individually bears almost all the inflow to this area.
\begin{table}[h]
\caption{Top 9 migration counts when city group integrated}
\begin{center}
\begin{threeparttable}
\begin{tabular}{l c}
    \textbf{Direction} & \textbf{Cnt.} \\ \hline
    Intra- Pearl River Delta & 1.00 \\
Intra- Yangtze River Delta & 0.52 \\
Intra- Beijing Area & 0.27 \\
Pearl River Delta - Yangtze River Delta & 0.22 \\
Yangtze River Delta - Pearl River Delta & 0.22 \\
Yangtze River Delta,Beijing Area & 0.21 \\
Beijing Area - Yangtze River Delta & 0.20 \\
Pearl River Delta - Beijing Area & 0.15 \\
Beijing Area - Pearl River Delta & 0.15 \\ \hline
\end{tabular}
\end{threeparttable}
\end{center}
\label{tab:area}
\end{table}

\subsection{Intra-city level migration}
When it comes to intra-city analysis, we study intra-district migration, as well as the process people moving from one community to another. We still use joined data of migration from September to December in 2015, data of this period is fresh and covers longer than bi-monthly comparison, meanwhile avoided summer holiday and winter holiday.

Table \ref{tab:inbj} shows top 5 net immigration destination district and top 5 net emigration origin district in Beijing. Haidian District attracted most people in the 3-month period, we are informed from the system that, although the government's policy encourages transition to southeast, to connect tighter with Tianjin and Hebei, whereas though people moving to the opposite direction. This, again, calls for consecutive monitoring to prove or to falsify.
\begin{table}[h]
\caption{Top 5 net immigration destination and top 5 net emigration origin by district in Beijing}
\begin{center}
\begin{threeparttable}
\begin{tabular}{l c}
    \textbf{District} & \textbf{Regularized migration count} \\ \hline
    HAIDIAN & 1.00 \\
CHAOYANG & 0.58 \\
FENGTAI & 0.21 \\
XICHENG & 0.20 \\
DONGCHENG & 0.10 \\
CHANGPING & -0.22 \\
SHIJINGSHAN & -0.24 \\
SHUNYI & -0.33 \\
DAXING & -0.35 \\
TONGZHOU & -0.72 \\ \hline
\end{tabular}
\end{threeparttable}
\end{center}
\label{tab:inbj}
\end{table}

In the last part of this section, there would be some cases on certain communities in Beijing, to see where the new inhabitants come from, and where the former residents leave for. It is noticeable that, in this scale, inter-city movings still counts for most of the migration, we select those happened with the capital city for better presentation of inter-communities moving processes.

\begin{figure}
\centerline{\includegraphics[width=8cm]{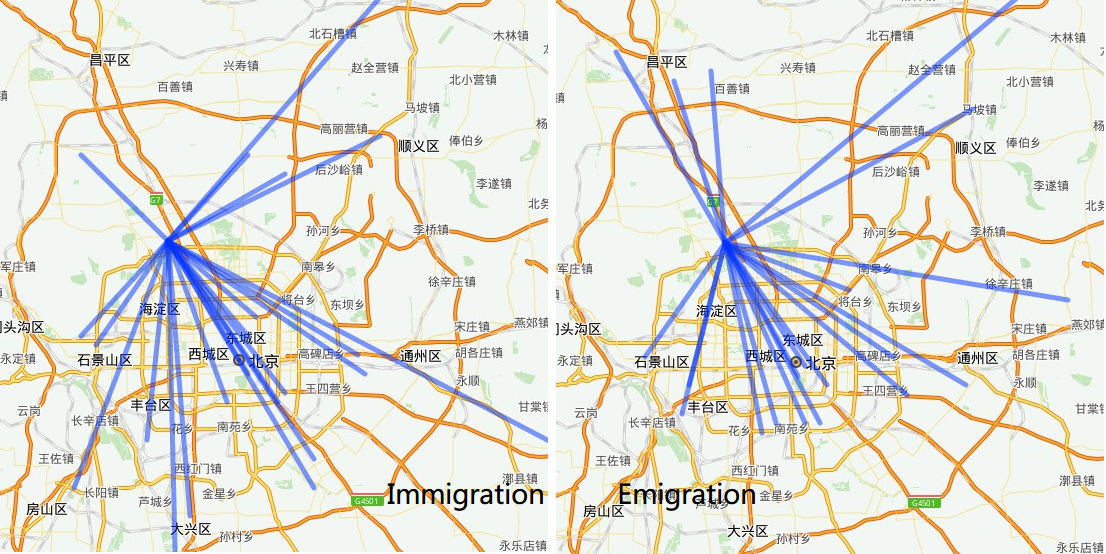}}
\caption{Immigration to/Emigration from Shangdi}
\label{fig:shangdi}
\end{figure}

Figure \ref{fig:shangdi} shows intra-city migration in which the destination or origin is Shangdi, a combination of three communities (Shangdi Dongli, Shangdi Xili, Shangdi Jiayuan) in Haidian district. It is relatively a new developed area, and many internet companies located not far from this place.

Some people moved from the same origin, some to the same destination, and totally there are more than thirty cases both in-bound and out-bound in this 3-month period. Only very few cases that the counterpart are at downtown, most of the counterpart places are rural places. Although both the two graphs are alike radiations, they are actually different with close examination: the center of gravity. The emigration destination is notably in the northwest of the immigration origin, it is same as the phenomenon shown in the large picture: people are moving from southeast to northwest.

\begin{figure}
\centerline{\includegraphics[width=8cm]{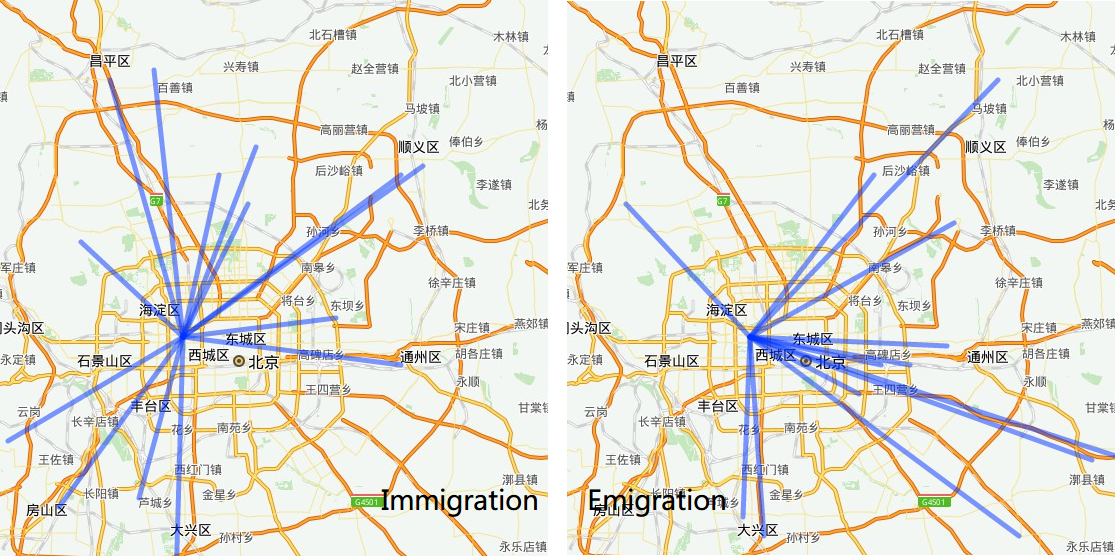}}
\caption{Immigration to/Emigration from Ganjiakou}
\label{fig:ganjiakou}
\end{figure}

Figure \ref{fig:ganjiakou} is another example. In this case our target is Ganjiakou, an old central area in downtown Beijing, some state ministries located here, along with many residential communities built in last century. The inbound and outbound transitions is typically in contrast with Shangdi in former example. Immigrations are from the northwest, and emigration are bound to southeast. This is in accordance with the government's action of moving the city administrations and municipal offices to southeast.

\section{Consecutive periodical migration observation}
Static data is better at demonstration and mostly are posterior knowledge. Consecutive data not only show dynamic changes, it also provides the evidence of predicting the future, there are many time series machine learning algorithms at hand for this purpose. In this case, if the target is to monitoring population migration, it would be a challenge for real time positioning generated transition data, since it is not able to distinguish business traveling, tourism and some other activities, it is individual based, whereas family play principle part in migration. Our system fits all the needs in such purpose.

\begin{figure}
\centerline{\includegraphics[width=8cm]{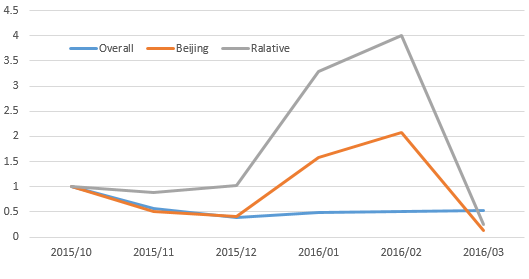}}
\caption{Net immigration in Beijing for consecutively 6 months}
\label{fig:period}
\end{figure}

In this paper we will take net immigration counts in Beijing as an example to illustrate consecutive monthly migration observation. In Figure \ref{fig:period}, the red line is monthly net immigration counts in Beijing, the baseline is net immigration in October 2015 in Beijing, the magnitude is ten thousand; the blue line is monthly total migration counts as a background, the purple line is the ratio of these two data source.

The sudden rise and fall in the chart happens to witness the last wave of real estate price hiking since year beginning of 2016, as well as the calm down since government regularization since March 2016. It is worth thinking why these two factors fits, no lagging and forecasting. Because as in common sense, there exists many real estate transection that are not for pure purpose of habitat. Actually with our data, a further study on this phenomenon is feasible soon.

\section{Conclusions and future work}
In this paper, we demonstrated a consecutively migration monitoring system from inter-province scale to intra-city scale. The data is updated in weekly basis in accumulation, several machine learning models are built and trained to identify population migration from noises. In the process of observing data drew from the system, many interesting phenomenon are witnessed. Some of the transitions fits our instinct, while there are also many cases against the policy makers' will. In macro-scale, with the help of this system, the government would be able to maker better regularization terms, plan better urban city, as well as to examine if the original purpose realized. In micro-scale of the society, the system helps in choosing better habitation, and make better investment choices.

Still, the power of the system is far from truly exploited, there are many respects for future work.

In machine learning aspect, prediction models would be the an effective way to help make decisions, deviations from prediction can also be exploited to find outliers of certain event. A classification model to identify if the habitats own the house or rent it would contribute features in real estate market related models.

In data processing aspect, further study on economic zone, not matter it is as small as combination of communities or as big as delta zone, rather than administrative divisions, should be done. The data can also be employed to examine those urban planning theories on target cities, for instance the shrinking city theory, the central place theory, etc..

\bibliographystyle{abbrv}

\end{document}